\documentclass[letterpaper, 10 pt, conference]{ieeeconf}  

\IEEEoverridecommandlockouts                              

\usepackage{amsmath}
\usepackage{amsthm,amsmath,amssymb}
\usepackage{mathrsfs}
\usepackage{bm}
\usepackage{algorithm}
\usepackage{algorithmic}
\usepackage{color}
\usepackage{graphicx}
\usepackage{makecell}
\usepackage{tabularx}
\usepackage{caption}
\usepackage{float} 
\usepackage{subcaption}
\usepackage{hyperref}

\title{\LARGE \bf
A Learning-based Adaptive Compliance Method for Symmetric Bi-manual Manipulation
}

\author{Yuxue Cao$^{1,*}$, Wenbo Zhao$^{2,*}$, Shengjie Wang$^{3,\dag}$, Xiang Zheng$^{4}$, \\Wenke Ma$^{5}$, Zhaolei Wang$^{6}$ and Tao Zhang$^{2,\dag}$ \textit {Fellow, IEEE}
\thanks{Our website can be found at \href{https://sites.google.com/view/lac-manipulation}{https://sites.google.com/view/lac-manipulation}.}
\thanks{$^{*}$Equal contribution. $^{\dag}$Corresponding author: wangsj23@mails.tsinghua. edu.cn, taozhang@tsinghua.edu.cn.}
\thanks{$^{1}$Beijing Institute of Control Engineering. $^{2}$Department of Automation, Tsinghua University. $^{3}$Institute for Interdisciplinary Information Sciences, Tsinghua University. $^{4}$Department of Computer Science, City University of Hong Kong. $^{5}$Qian Xuesen Laboratory of Space Technology. $^{6}$Beijing Aerospace Automatic Control Institute.}%
}

\begin{document}

\maketitle
\thispagestyle{empty}
\pagestyle{empty}

\begin{abstract}

Symmetric bi-manual manipulation is an essential skill in on-orbit operations due to its potent load capacity. Previous works have applied compliant control to maintain the stability of manipulations. However, traditional methods have viewed motion planning and compliant control as two separate modules, which can lead to conflicts with the simultaneous change of the desired trajectory and impedance parameters in the presence of external forces and disturbances. Additionally, the joint usage of these two modules requires experts to manually adjust parameters.
To achieve high efficiency while enhancing adaptability, we propose a novel Learning-based Adaptive Compliance algorithm (LAC) that improves the efficiency and robustness of symmetric bi-manual manipulation. Specifically, the algorithm framework integrates desired trajectory generation and impedance-parameter adjustment under a unified framework to mitigate contradictions and improve efficiency. Second, we introduce a centralized Actor-Critic framework with LSTM networks preprocessing the force states, enhancing the synchronization of bi-manual manipulation. When evaluated in dual-arm peg-in-hole assembly experiments, our method outperforms baseline algorithms in terms of optimality and robustness.

\end{abstract}

\section{INTRODUCTION}

Planning and control of space manipulators play an essential role in on-orbit services, such as assembling and maintaining large space facilities, refueling various spacecraft, and establishing extraterrestrial bases \cite{jiang2022progress}. 
Symmetric bi-manual manipulation is credible for handling objects and performing the peg-in-hole assembly operation, which is essential for the on-orbit assembly of large space facilities or the construction of extraterrestrial bases. Nevertheless, it faces challenges such as inaccurate dynamic models and unsynchronized control of manipulators \cite{wu2020reinforcement}. 
Additionally, the environmental contact force interacting with the object must also be considered during manipulation. Therefore, current research commonly combines motion planning and compliance control to achieve the task \cite{1998Synthesis}, whereas most of them perform motion planning first and then track the desired trajectory with impedance control. The separate control strategy is inefficient, and the simultaneous changes to these two modules may cause conflicts under certain circumstances. 
Furthermore, to enhance the robustness of impedance control, some researchers have proposed various optimization methods for selecting impedance parameters online \cite{8593853,2021Coordinated,20201908638489}
, while previous methods need prior information and extra assumptions. 
Therefore, it remains an open challenge to design a synchronous framework that contains adaptive compliance and motion planning.

In recent years, model-free reinforcement learning (RL) has shown promising results in robot manipulation tasks \cite{ju2022transferring}. We can leverage RL methods to enable a unified framework for motion planning and parameter adjustment tasks, and optimize the planning policy and impedance parameters through interaction experiences. In addition, specific contact dynamics models are unnecessary for model-free methods, eliminating the need for prior information. 
Compared with single-arm manipulation \cite{9830834, 9361338}, symmetric bi-manual manipulation requires higher synchronization of dual-arm control, which is particularly important when implementing the cooperative peg-in-hole assembly task. In conclusion, RL-based methods can facilitate symmetric bi-manual manipulation to achieve high efficiency, robustness, and synchronization.

In this paper, we propose the Learning-based Adaptive Compliance (\textbf{LAC}) algorithm for symmetric bi-manual manipulation. 
The proposed method addresses the synchronous update of both motion planning and compliant control for bi-manual manipulation. 
Furthermore, it can complete the dual-arm cooperative peg-in-hole assembly operation with a 2mm clearance in which the two arms hold the same peg. The contributions of this paper are as follows:
\begin{itemize}
    \item We develop a novel algorithm framework to complete symmetric bi-manual manipulation based on RL. The LAC generates the desired trajectory of dual-arm operations, improving motion planning efficiency. Meanwhile, it adaptively updates impedance parameters to overcome the poor robustness of impedance control.
    
    \item We design a centralized Actor-Critic network structure to achieve cooperative planning and impedance-parameters modification. The centralization of the framework is useful in operations requiring high dual-arm synchronization. Moreover, results demonstrate that the force trend feature processed by our network greatly improves the performance of compliance operations.
    
    \item We verify our method on a typical symmetric bi-manual manipulation task, the dual-arm cooperative peg-in-hole assembly operation. In both simulation environments and real-world settings, the results further validate the feasibility and robustness of the proposed algorithm.
\end{itemize}


\section{RELATED WORK}
This section reviews the literature on algorithms that aim to accomplish symmetric bi-manual manipulation.

\subsection{Traditional Methods}

In the dual-arm cooperative handling operation, researchers such as Yan \cite{8593853}, Hu \cite{2021Coordinated}, and Liu \cite{s21144653} utilized impedance control to manage the manipulators' operating forces. 
Caccavale proposed a double-loop impedance control method based on the above ideas \cite{4639601}. In this method, the centralized impedance controller facilitates compliant contact between the object and the environment, while the decentralized impedance controller controls the force between the end-effectors and the object. A host of researchers, including Heck\cite{20140717299306} and Tarbouriech\cite{20201908638489}, further refined the compliant control framework for symmetric bi-manual manipulation based on \cite{4639601}. 
Most of the studies cited above utilized manually calibrated impedance parameters, which cannot adapt to changes in the system or environment. 
Recently, several algorithms that can adjust impedance parameters online have been proposed, such as those based on nonlinear optimization \cite{8273239}, quadratic optimization \cite{8593853}, adaptive law compensation \cite{9254995}, and neural network \cite{8812496}. 
While these methods consider the control of internal and external forces, it is currently only applied to dual-arm cooperative handling operations. Though there are several studies on the single-arm peg-in-hole assembly \cite{WOS:000778988400068, WOS:000543336600003}
, research on dual-arm assemblies is scarce and mainly involves unsymmetrical cooperative operations, where one arm holds the peg and the other holds the hole \cite{peg-in-hole2020yanjiang}. To our knowledge, there is no published research on the symmetrical dual-arm peg-in-hole assembly operation, where both arms hold the same peg. 


\subsection{Reinforcement Learning Methods}

Model-free reinforcement learning was initially used for single-arm compliant operations, where Stulp et al. used the $PI^2$ algorithm to learn variable impedance, adapting to both deterministic and stochastic force fields \cite{6227337}. With the development of neural networks, 
Beltran-Hernandez applied reinforcement learning to single-arm peg-in-hole assembly, which utilized TCN and MLP networks to pre-process observations \cite{20200595648}. Several references, including \cite{8967946,20191206663241,8868579,9361338}, 
focused on adjusting impedance parameters to reduce the contact force between the peg and the hole and the assembly time. While reinforcement learning has made remarkable achievements in single-arm operations, few studies have focused on dual-arm operations. Alles and Aljalbout proposed a reinforcement learning framework with a centralized policy network and two decentralized single-arm controllers \cite{20210361614}. Its effectiveness was verified in the asymmetric peg-in-hole assembly. However, as far as we know, no study has yielded the effect of reinforcement learning in symmetric bi-manual manipulation with strong kinematics and dynamics constraints. The closed-chain system requires higher synchronization of dual-arm control, which brings new challenges to the application of reinforcement learning. 

\section{PROBLEM FORMULATION}


\subsection{Kinematic and Dynamic Model of the Dual-Arm Closed-Chain System}
\subsubsection{Kinematic Model}

\begin{figure}[!t]
  \centering
  \includegraphics[width=0.85\hsize]{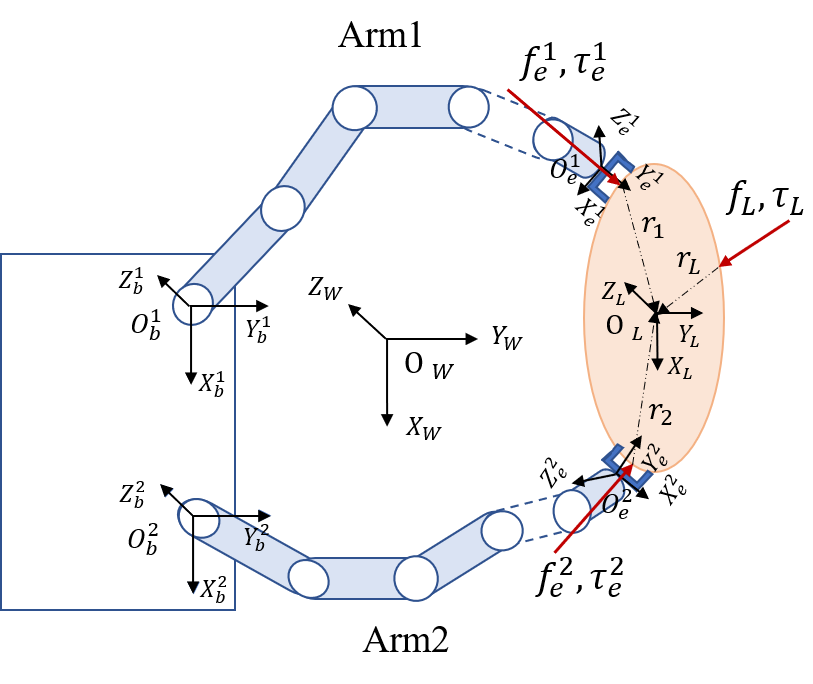}
  \caption{Closed-chain model of the dual-arm robot.}
  \label{KDmodel}
\end{figure}

To maintain a stable connection between the end-effectors and the object, the whole movement process must comply with the closed-chain kinematics constraints formulated as follows:
\begin{equation}
\label{eq1}
{T^{{b}^i}_{{e}^i}}={T^{{b}^i}_{W}}T^{W}_{L}{T^L_{{e}^i}}
\end{equation}
where $T^B_A$ represents the transformation matrix of the coordinate system $A$ relative to the coordinate system $B$. 

\subsubsection{Dynamic model}
In Fig. \ref{KDmodel}, $f_L$ and $\tau_L$ are the external force and moment exerted on the object by the environment. $f_{e}^i$ and $\tau_{e}^i$ are the force and moment applied to the object by the i-th manipulator. $r_L$, $r_1$, $r_2$ are vectors from the force points of $f_L$, $f_{e}^1$, $f_{e}^2$ to the object's centroid. $v_L$ and $w_L$ are the velocity and angular velocity of the object. $M_L$, $I_L$ and $G_L$ are the object's mass, moment of inertia and gravity, respectively. The dynamic model of the dual-arm closed-chain system can be formulated as:
\begin{equation}
\label{eq2}
\begin{split}
\left[\begin{matrix}I_3&O\\\left(r_1\right)^\ast&I_3\\\end{matrix}\right]\left[\begin{matrix}f_{e}^1\\\tau_{e}^1\\\end{matrix}\right]+\left[\begin{matrix}I_3&O\\\left(r_2\right)^\ast&I_3\\\end{matrix}\right]\left[\begin{matrix}f_{e}^2\\\tau_{e}^2\\\end{matrix}\right]+\left[\begin{matrix}G_L\\O\\\end{matrix}\right]\\+\left[\begin{matrix}I_3&O\\\left(r_L\right)^\ast&I_3\\\end{matrix}\right]\left[\begin{matrix}f_L\\\tau_L\\\end{matrix}\right]=\left[\begin{matrix}m_L\dot{v}_L\\I_L{\dot{\omega}}_L+\omega_L\times\left(I_L\omega_L\right)\\\end{matrix}\right]
\end{split}
\end{equation}
where $(\cdot)^\ast$ represents the transpose of the vector. Eq. \ref{eq2} can be simplified as $-\Gamma_1F_{e}^1-\Gamma_2F_{e}^2=-{\overline{G}}_L-\Gamma_LF_L+F_{IL}$, where the generalized force $F$ represents the vector composed of force and moment. 
Since the motion planning algorithms generate the desired trajectory for the object, and in turn determines $F_{IL}$, the primary challenge of this problem is to decompose that of the object's centroid into the desired generalized forces of the end-effectors. This paper refers to the shared force mode proposed in \cite{yan2016coordinated}, obtaining the desired generalized forces of the end-effectors as follows:
\begin{equation}
\label{eq4}
\left[\begin{matrix}F_{e}^1\\F_{e}^2\\\end{matrix}\right]=\left[\begin{matrix}-\Gamma_1&-\Gamma_2\\\end{matrix}\right]^{\#}\left(-{\overline{G}}_L-\Gamma_L F_L+F_{IL}\right)
\end{equation}
where $\#$ represents the pseudo inverse of the matrix.

\subsection{Position-based Impedance Control}
The position-based impedance control converts force errors into position adjustment, and realizes force control by adjusting the desired trajectory. The second-order dynamic model between force and position is:
\begin{equation}
\label{eq5}
M_d{(\ddot{X}-\ddot{X}}_d)+B_d(X-{\dot{X}}_d)+K_d(X-X_d)=F-F_d
\end{equation}
where $M_d$, $B_d$ and $K_d$ are the positive definite matrices of desired inertia, damping and stiffness of the impedance model, respectively. Diagonal matrices is usually selected to obtain the linear decoupling response. $\ddot{X}$, $\dot{X}$, $X$ represent the actual acceleration, velocity and position, respectively. ${\ddot{X}}_d$, ${\dot{X}}_d$, $X_d$ are the desired acceleration, velocity and position. $F_d$ and $F$ are the desired and the actual contact force, respectively.

The Lagrangian transformation and discretization are applied to Eq. \ref{eq5} to deduce position adjustment from force errors, as Eq. \ref{eq6} shows.
\begin{equation}
\begin{split}
\label{eq6}
\delta X\left(t\right)=\frac{T^2}{\omega_1}\left(E\left(t\right)+2E\left(t-1\right)+E\left(t-2\right)\right)-\\\frac{\omega_2}{\omega_1}\delta X\left(t-1\right)-\frac{\omega_3}{\omega_1}\delta X\left(t-2\right)
\end{split}
\end{equation}
where $E(t)=F(t)-F_d(t)$, and $T$ is the sampling period. $\omega_1=4M_d+2B_dT+K_dT^2$, $\omega_2=-8M_d+2K_dT^2$, and $\omega_3=4M_d-2B_dT+K_dT^2$. Then the actual trajectory at time $t$ can be calculated as: $X\left(t\right)=X_d\left(t\right)+\delta X\left(t\right)$. 

\begin{figure*}
  \centering
  \includegraphics[width=\hsize]{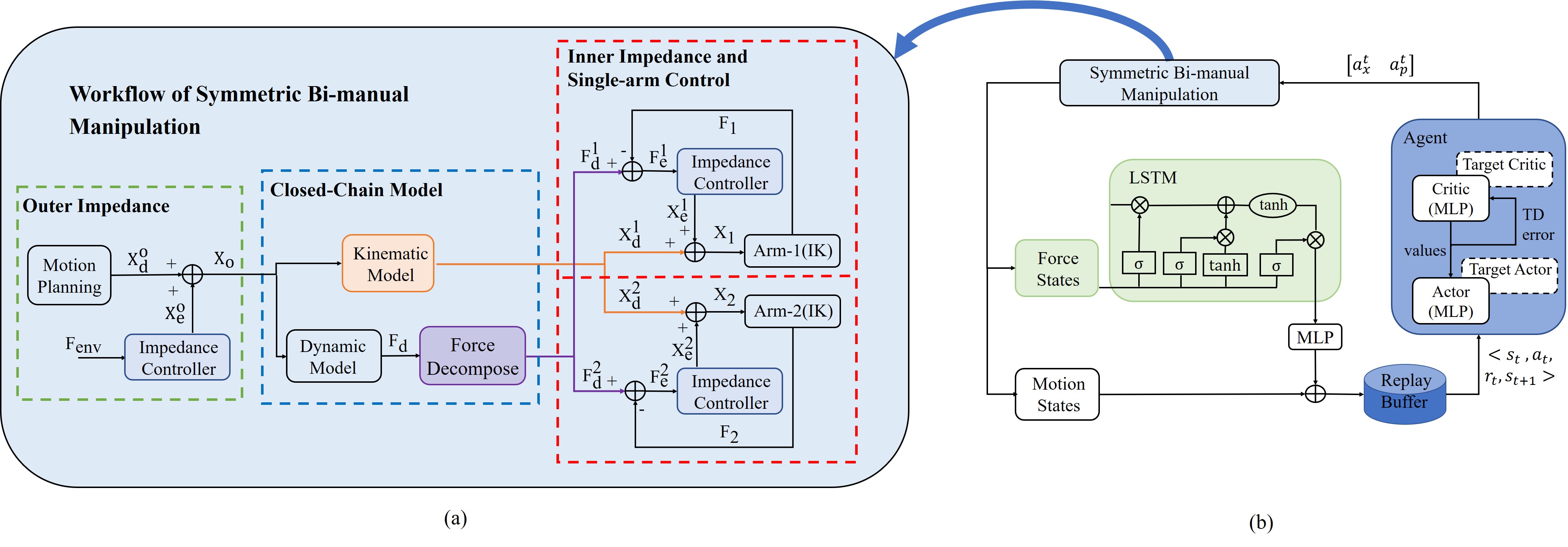}
  \caption{Algorithm framework of \textbf{LAC}. (a) Workflow of symmetric bi-manual manipulation consists of three parts. (b) Adaptive compliance algorithm based on reinforcement learning.}
  \label{Framework}
\end{figure*}

\section{ADAPTIVE COMPLIANCE ALGORITHM BASED ON REINFORCEMENT LEARNING}

\subsection{Reinforcement Learning}
The process of reinforcement learning can be described as a Markov decision process (MDP). The MDP process comprises five components: $<S,\ A,\ R,\ P,\ \rho_0>$. The agent gets the environment state $s_t \in S$ and generates the action $a_t \in A$ using the current policy $\pi(s_t)$. Once $a_t$ is performed, the environment transitions to a new state $s_{t+1}$, and the agent obtains a reward $r_t$. Reinforcement learning endeavors to find the optimal policy $\pi^*$ that yields the maximum expected reward value. For this paper, we utilize the Soft Actor-Critic (SAC) algorithm \cite{haarnoja2018soft}. Unlike other deep reinforcement learning algorithms, SAC takes into account both the reward value and policy entropy maximization. Maximizing policy entropy facilitates the exploration of policy, thereby allowing the algorithm to find the optimal solution more effectively in high-dimensional optimization problems. 

\subsection{Workflow of Symmetric Bi-manual Manipulation}

In this paper, we extend the double-loop impedance method proposed in \cite{4639601} and adapt it to symmetric bi-manual manipulation. The manipulation workflow is illustrated in Fig. \ref{Framework}, which comprises three parts. After the motion planning model generates the object's desired trajectory, denoted as $X_{d}^o$, the outer impedance adjusts the object's position to minimize the external force, denoted as $F_{env}$, which the environment exerts on the object. The second-order impedance model can be rewritten as Eq. \ref{eq8} for the outer impedance.

\begin{equation}
\label{eq8}
M_{d}^o{\ddot{X}}_{e}^o+B_{d}^o{\dot{X}}_{e}^o+K_{d}^oX_{e}^o=F_{env}
\end{equation}
where $X_{e}^o=X_o-X_{d}^o$ is the adjustment to the object's desired trajectory $X_{d}^o$.  After obtaining $X_e^o$ through Eq. \ref{eq8}, we can compute the compliant object trajectory $X_o$ by adding $X_e^o$ and $X_d^o$. The complaint desired trajectory $X_o$ decreases the interactive environmental force. 

The second part pertains to the dual-arm closed-chain kinematics and dynamics model. This part calculates the desired trajectory $X_{d}^i$ and the desired force $F_{d}^i$ of the manipulator according to Eq. \ref{eq1} and Eq. \ref{eq4}, respectively. 

Once $X_{d}^i$ and $F_{d}^i$ are calculated, the inner impedance and single-arm control are implemented to drive the motion of manipulators. The inner loop's second-order impedance model is formulated as:
\begin{equation}
\label{eq9}
M_{d}^i{\ddot{X}}_{e}^i+B_{d}^i{\dot{X}}_{e}^i+K_{d}^iX_{e}^i=F_{e}^i
\end{equation}
where $X_{e}^i=X_i-X_{d}^i$ is the adjustment to the desired trajectory of the i-th manipulator's end-effector $X_{d}^i$. $F_{e}^i=F_{d}^i-F_i$ is the error between the end-effector's desired force and the actual contact force.

\subsection{Learning-based Adaptive Compliance(\textbf{LAC}) Algorithm}
We design an adaptive compliance algorithm based on reinforcement learning, which aims to enhance the efficiency of motion planning and bi-manual manipulation compliance.
\subsubsection{Algorithm Framework}
As shown in Fig. \ref{Framework}(b), the framework of our proposed learning-based adaptive compliance method for symmetric bi-manual manipulation is a centralized framework comprising two components. The high-level module, which runs reinforcement learning at a fixed frequency of 20 Hz, provides the object's desired trajectory and impedance parameters. Additionally, it acquires the states of two manipulators simultaneously. On the other hand, the low-level module involves impedance control operating at a frequency much higher than the high-level module. The slow control frequency of high-level reinforcement learning allows the agent to process environmental states and generate the next action, while the high frequency of the low-level module ensures precise and prompt control of the manipulators.


The reinforcement learning policy outputs $\left\{a_x,a_p\right\}$, where $a_x=[\Delta x, \Delta y, \Delta z]\in R^3$ corresponds to the object's position change and represents the motion planning module, and $a_p=[B_d, K_d] \in R^6$ corresponds to the desired damping matrix and stiffness matrix diagonal values of the impedance controller. The desired inertia matrix remains constant because it significantly affects the system's stability. Furthermore, to maintain the control system's stability, the impedance parameters' upper and lower limits are defined as hyper-parameters of the algorithm, $a_{p}^{max}$ and $a_{p}^{min}$, respectively. In the low-level module, the impedance controller with variable parameters modifies the desired trajectory $X_{d}$ output by the policy, resulting in compliant operations.


\subsubsection{Neural Network Architecture}
As the environmental states $s_t$ are divided into motion states $m_t$ and force states $f_t$, the manipulator's motion constraints limit the change rate of $m_t$, and the states between two moments have strong continuity. Nevertheless, $f_t$ are related to current contact dynamics and hence show intense volatility. Moreover, adjusting the sub-goal position and impedance parameters should consider the current force information and its trend. Therefore, we enable the agent to obtain the force states at the low-level frequency to enhance the states' continuity and promote better algorithm performance. 
Since Long Short-Term Memory (LSTM) establishes connections between nodes of different hidden layers and can use past moments' information to infer the future moment's state, we utilize LSTM networks to the original deep reinforcement learning MLP network architecture, intending to process the force states sequence. 
A fully connected layer then processes the output of the LSTM networks to obtain the feature vector of the force states. After that, the motion states are connected with the pre-processed force feature vector, which is then input into the agent's neural networks. The algorithm pseudo-code of our approach is available in Algorithm \ref{alg:1}.

\subsection{Dual-arm Cooperative Peg-in-hole Assembly}

\begin{figure}[!t]
  \centering
  \includegraphics[width=\hsize]{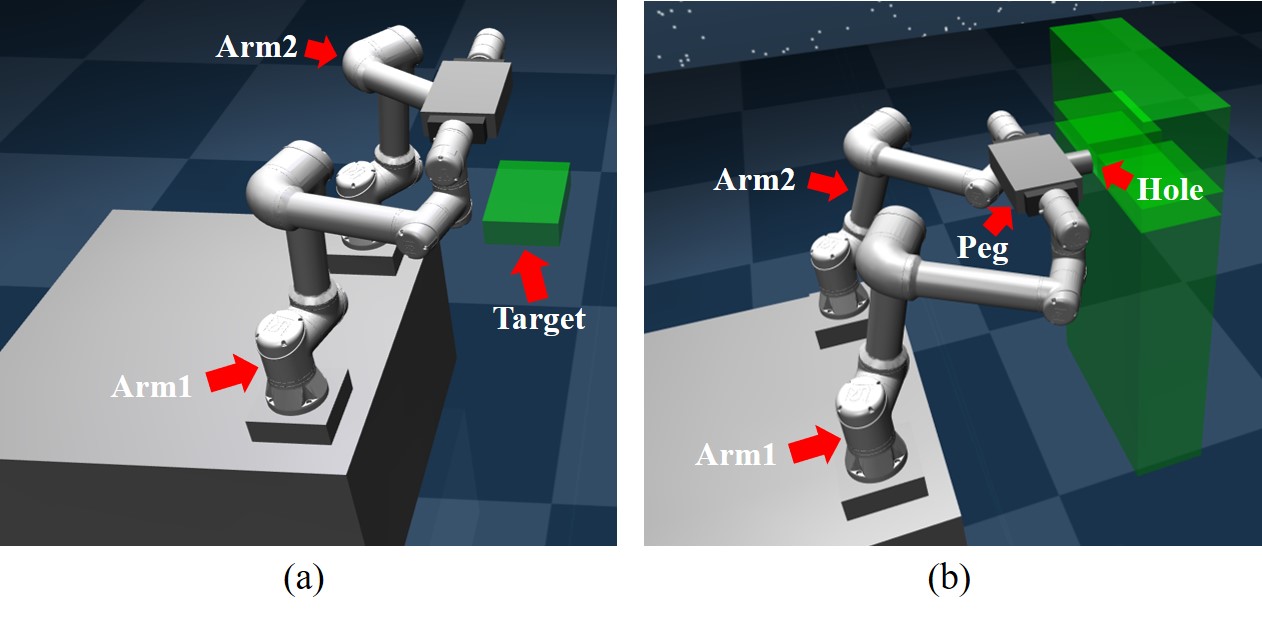}
  \caption{The two stages of the dual-arm cooperative peg-in-hole assembly task. The object needs to be first transported to the target position and then slowly inserted.}
  \label{SimEnv}
\end{figure}

Fig. \ref{SimEnv} depicts the simulation environment of the dual-arm cooperative peg-in-hole assembly operation. 
The agent's observations include end-effectors' poses $P_{e}^{1,2},\Phi_{e}^{1,2}\in R^6$, velocities $v_{e}^{1,2},\omega_{e}^{1,2}\in R^6$, and the pose of the object's centroid $P_o,\Phi_o\in R^6$. Additionally, the error vector between the current position and the target position of the object is included in the state space to enhance learning efficiency. Note that the algorithm primarily focuses on the interaction states between the peg and hole during assembly, and as a result, the state space omits the manipulators' joint information. Additionally, the state space includes the contact force between the peg and the hole ($F_{env}\in R^3$) and the peg insertion depth ($d$). The policy outputs the object's centroid positional variation $[\Delta x, \Delta y, \Delta z]$, as well as the desired damping $B_{d}^o\in R^3$ and stiffness $K_{d}^o\in R^3$ of the outer impedance controller. Furhtermore, to improve learning efficiency in dual-arm assembly, we adopt a hybrid policy that leverages the knowledge of the PD position error controller. The object's desired trajectory ($X_{d}^o$) is the sum of the PD position controller's output ($X_g$) and the reinforcement learning policy's output ($a_x$), expressed as $X_{d}^o=X_g+a_x$. The hybrid policy accelerates the search process before the peg contacts the hole. 

As for the reward function, the peg’s insertion depth is the primary indicator of successful assembly. Additionally, minimizing the interaction force between the peg and hole reduces wear between parts during insertion. However, to prevent the agent from avoiding insertion, we decrease the weight of contact force in the reward function. Completion time is another performance indicator. The reward value increases as the completion time shortens, encouraging the agent to complete the task quickly. Therefore, the reward function is structured as follows. 
\begin{equation}
\label{eq12}
r = \left\{ 
\begin{aligned}
&1.05-tanh(d)-0.05 \times tanh(F_{env}), \enspace F_{env} \leq F_{max} \\
&100+((1-t/T) \times 100), \enspace d \geq d_{s} \\
&-10, \enspace          Not \enspace safe 
\end{aligned}
\right.
\end{equation}
where $d_{s}$ is the target insertion depth. 
When safety problems occur, we will penalize the agent with -10 and terminate the current episode. The safety problems include two aspects: 1) the interaction force between the peg and the hole exceeds the defined threshold $F_{max}$; 2) the closed-chained constraint is not satisfied during the assembly process, causing the object to fall away.

\section{SIMULATION RESULTS AND REAL WORLD EXPERIMENTS}
\subsection{Simulation Environment and Real-World Settings} \label{serws}

\begin{figure}[t]
  \centering
  \includegraphics[width=\hsize]{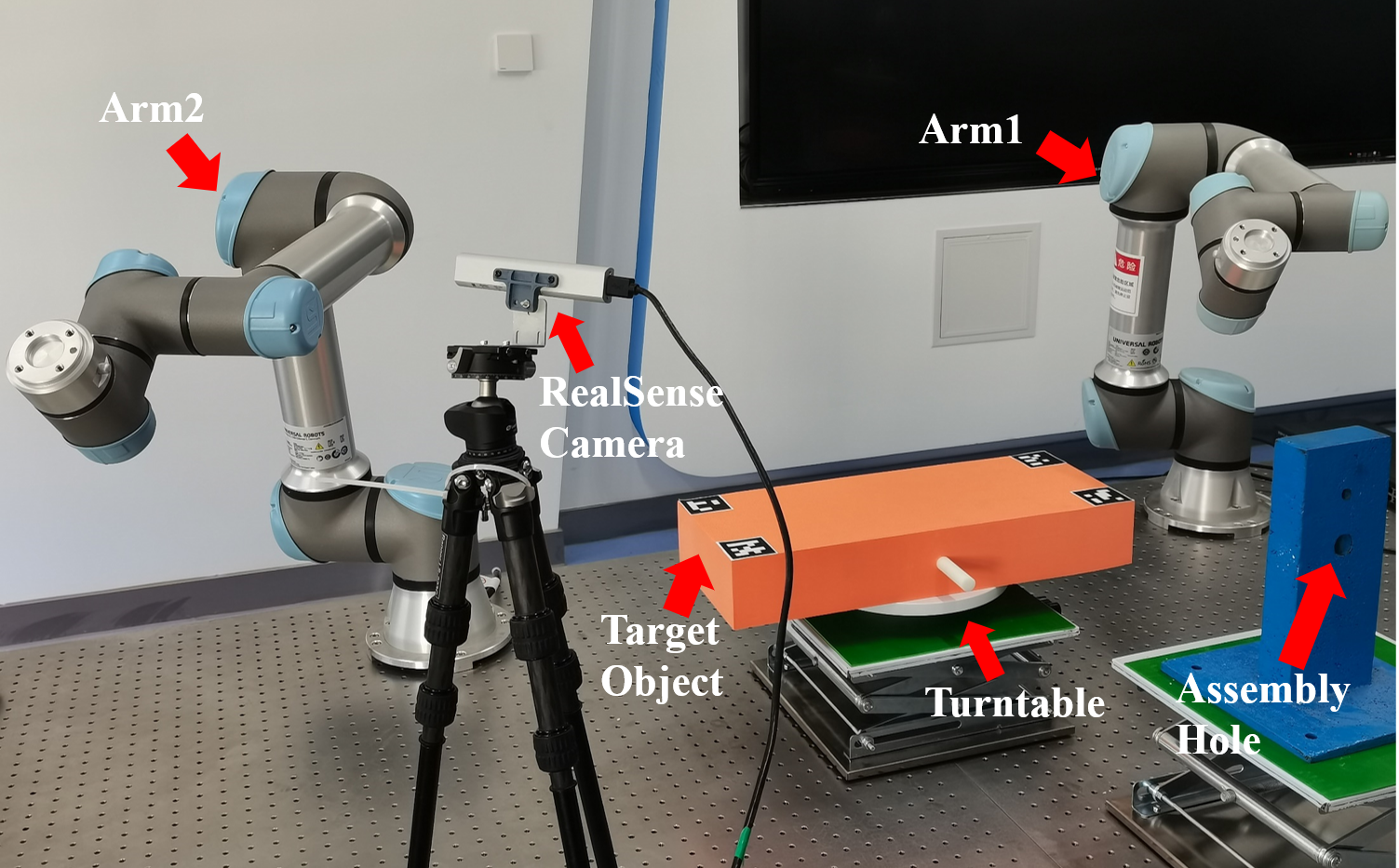}
  \caption{The ground-based dual-arm robot system for typical on-orbit assembly operation.}
  \label{RealWorld}
\end{figure}

\begin{table}[t]
\begin{center}
\caption{Hyperparameters of \textbf{LAC}}
\label{hp of LAC}
\begin{tabular}{|c|c|}
\hline
Hyperparameters& LAC\\
\hline
LSTM/MLP network & 16/3 \\
Actor network & (256,256) \\
Critic network & (256,256) \\
Learning rate of actor & 1.e-3  \\
Learning rate of critic & 5.e-4  \\
Optimizer & Adam \\
ReplayBuffer size& $10^6$ \\
Discount ($\gamma$)& 0.995 \\
Polyak ($1-\tau$)& 0.995 \\
Batch size & 128 \\
Length of an episode & 200 steps \\
Maximum steps & 1e6 steps \\
\hline 
\end{tabular}
\end{center}
\end{table}

\begin{algorithm}[t!]
	\renewcommand{\algorithmicrequire}{\textbf{Input:}}
	\renewcommand{\algorithmicensure}{\textbf{Output:}}
	\caption{Learning-based Adaptive Compliance(LAC)}
	\label{alg:1}
	\begin{algorithmic}[1]
	    \STATE Randomly initialize the parameters of the state pre-process networks (including LSTM, MLP); randomly initialize the agent's actor network with $l_{a}, w_{a}, \phi$ and critic network with $l_{c}^i,w_c^i,\theta_{i}$ , where $i=1,2$
	    \STATE Initialize replay buffer $\mathcal D$, the frequency of the high-level module $h_{h}$ and low-level module $h_{l}$
	    \STATE Initialize the parameters of target network with ${l^i_{c}}^{\prime}, {w^i_{c}}^{\prime}, \theta^\prime_{i} \leftarrow l_{c}^i, w_{c}^i, \theta_{i}, i=1,2$
	    \FOR {episode $e = 1, ... , E $}
	     \STATE Sample initial state $s_0$ 
	    \FOR {step $t=0, ... , T-1$} 
	    \STATE Divide $s_t$ into $m_t$ and $f_t$, input $f_t$ into LSTM networks to get the force feature vector $v_t$
	    \STATE Combine $m_t$ with $v_t$ to obtain the feature vector $c_t$
	    \STATE Sample an action $a_t=\left\{a_{x}^t,a_{p}^t\right\}$ from $\pi_\phi(a_t|c_t)$
            \FOR {step $u=0, h_{l}/h_{h}-1$}
            \IF {outer impedance}
	    \STATE Set outer impedance parameters to be $a_{p}^t$, and use $a_{x}^t$ to calculate $X_o$ according to Eq.\ref{eq8}
            \STATE Use $X_o$ to calculate $X_{d}^i$ and $F_{d}^i$ of the end-effectors according to Eq.\ref{eq1} and Eq.\ref{eq4}
            \ELSIF{inner impedance}
            \STATE Use $a_{x}^t$ to calculate $X_{d}^i$ and $F_{d}^i$ of the end-effectors according to Eq.\ref{eq1} and Eq.\ref{eq4}
            \STATE Set inner impedance parameters to be $a_{p}^t$
            \ENDIF
	    \STATE Calculate $X_i$ according to Eq.\ref{eq9}, and convert it to the desired joint angles ${q_{d,i}^u}$ to control the manipulator
            \ENDFOR
	    \STATE observe a new state $s_{t+1}$ and store $<s_t,a_t,r_t,s_{t+1}>$ into $\mathcal{D}$
	    \ENDFOR
	    \FOR{iteration $n = 1, N$}
	    \STATE Sample a minibatch $\mathcal B$ from replay buffer $\mathcal D$
	    \STATE Update critic network $l_{c}^i, w_{c}^i, \theta_{i}, i=1,2$ using minibatch $\mathcal B$
	    \STATE Update actor network $l_{a}, w_{a}, \phi$ using minibatch $\mathcal B$
	    \STATE Soft update the parameters of target networks, ${l^i_{c}}^\prime, {w^i_{c}}^\prime, \theta^\prime_{i}, i=1,2$
	    \ENDFOR
	    \ENDFOR
	\end{algorithmic}
\end{algorithm}

We first build simulation environments for the dual-arm cooperative peg-in-hole assembly task using Mujoco, a widely used simulation platform in reinforcement learning. The target position of the object's centroid is located in a $[0.1m, 0.3m, 0.12m]$ cube. The hyperparameters of \textbf{LAC} are illustrated in Table \ref{hp of LAC}.

\begin{figure*}[!t]
  \centering
  \includegraphics[width=0.9\hsize]{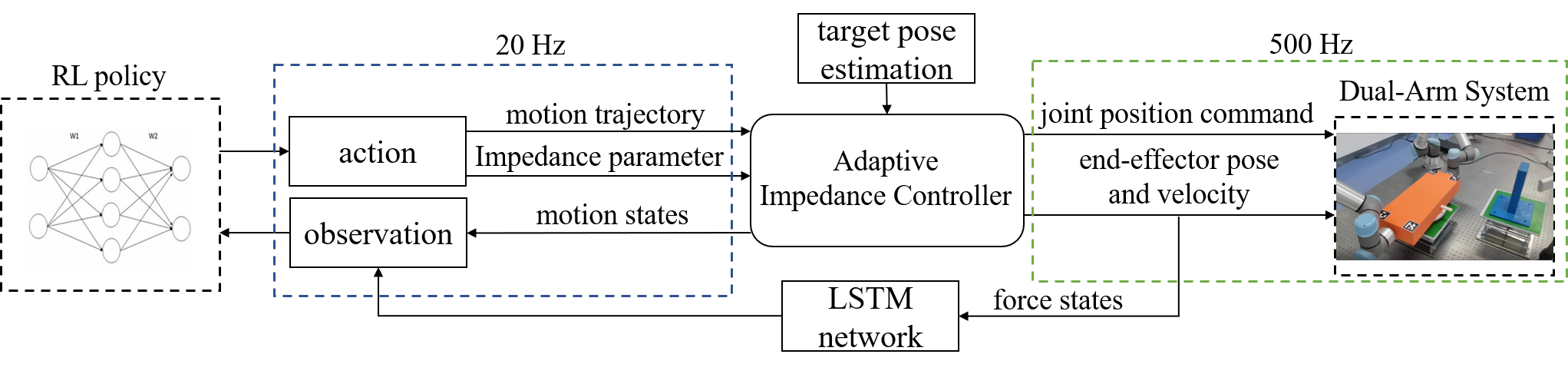}
  \caption{Block diagram of experimental system of adaptive compliance control algorithm based on reinforcement learning.}
  \label{FrameW}
\end{figure*}

To further validate the proposed algorithm, we construct a ground-based dual-arm robot system for typical on-orbit assembly operations, as shown in Fig. \ref{RealWorld} . The whole system mainly consists of two six-DoF UR robot arms, a target object, a turntable, an assembly hole, a RealSense camera and a laptop computer. The laptop  serves as the master computer for the two UR robot arms, running the dual-arm cooperative motion planning algorithm and the compliant 
control algorithm and sending the desired joint angles to the joint controllers of each arm via TCP/IP protocol. During the experiment, 
the data required by the master computer is estimated by the RealSense camera, which are used as inputs of a pre-trained reinforcement learning policy network.
The framework of the real-world system 
is shown in Fig. \ref{FrameW} .

We argue that our algorithm's performance improvement depends on four key factors: using SAC as reinforcement learning algorithm which takes policy entropy into account, introducing impedance mechanism into the closed-chain dual-arm system, variable impedance parameters, and pre-processing force states with LSTM networks. 
We provide the abbreviation of each algorithm: the algorithm with \textbf{TD3}, the algorithm only using RL training without impedance controller (\textbf{SAC}), the algorithm with fixed impedance parameters (\textbf{LAC w/ fixed Imp}) and the algorithm without LSTM networks (\textbf{LAC w/o LSTM}).

\subsection{Dual-arm Cooperative Assembly}

\subsubsection{Ablation Studies}

\begin{figure}[!t]
  \centering
  \includegraphics[width=\hsize]{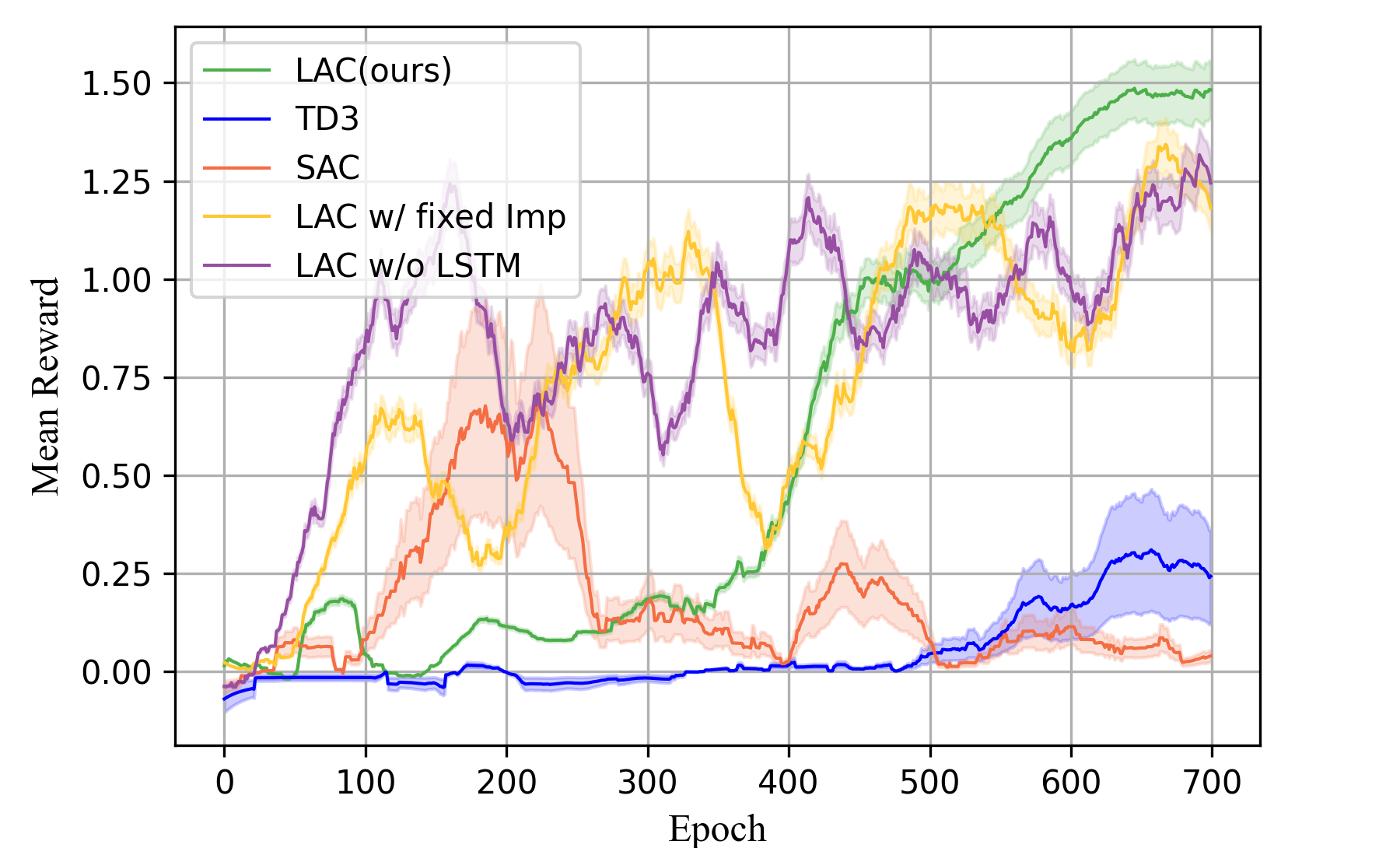}
  \caption{Mean reward curve during the training of different algorithms for the dual-arm peg-in-hole assembly operation (Curves smoothed report the mean of 3 times, and shadow areas show the variance.).}
  \label{PegReward}
\end{figure}

Fig. \ref{PegReward} depicts the average reward curve during the training of different algorithms under the dual-arm cooperative assembly scenario. The shaded regions correspond to the standard deviation using three different randam seeds. Our algorithm displays a more stable and larger convergence value compared to the others. The low reward value of \textbf{TD3} implies that it is hard to find a satisfactory policy for the assembly task without enough exploration. The reward value of \textbf{SAC} increases initially but declines rapidly and ultimately maintains a low value, which indicates that the interaction force between the peg and the hole exceeds the acceptable range without outer impedance and the training process fails. As \textbf{LAC w/ fixed Imp} has a lower action space dimension with only the position change outputs, and \textbf{LAC w/o LSTM} does not need to learn the parameters of LSTM networks, the reward values of these two algorithms rise more rapidly than \textbf{LAC} at the initial stage. However, \textbf{LAC w/ fixed Imp} cannot real-time adjust the impedance relationship between the peg and the hole, and \textbf{LAC w/o LSTM} only considers the current force rather than the trend of force change. Thus, their average reward values fluctuate violently during the training.  

\begin{figure}[!t]
  \centering
  \includegraphics[width=\hsize]{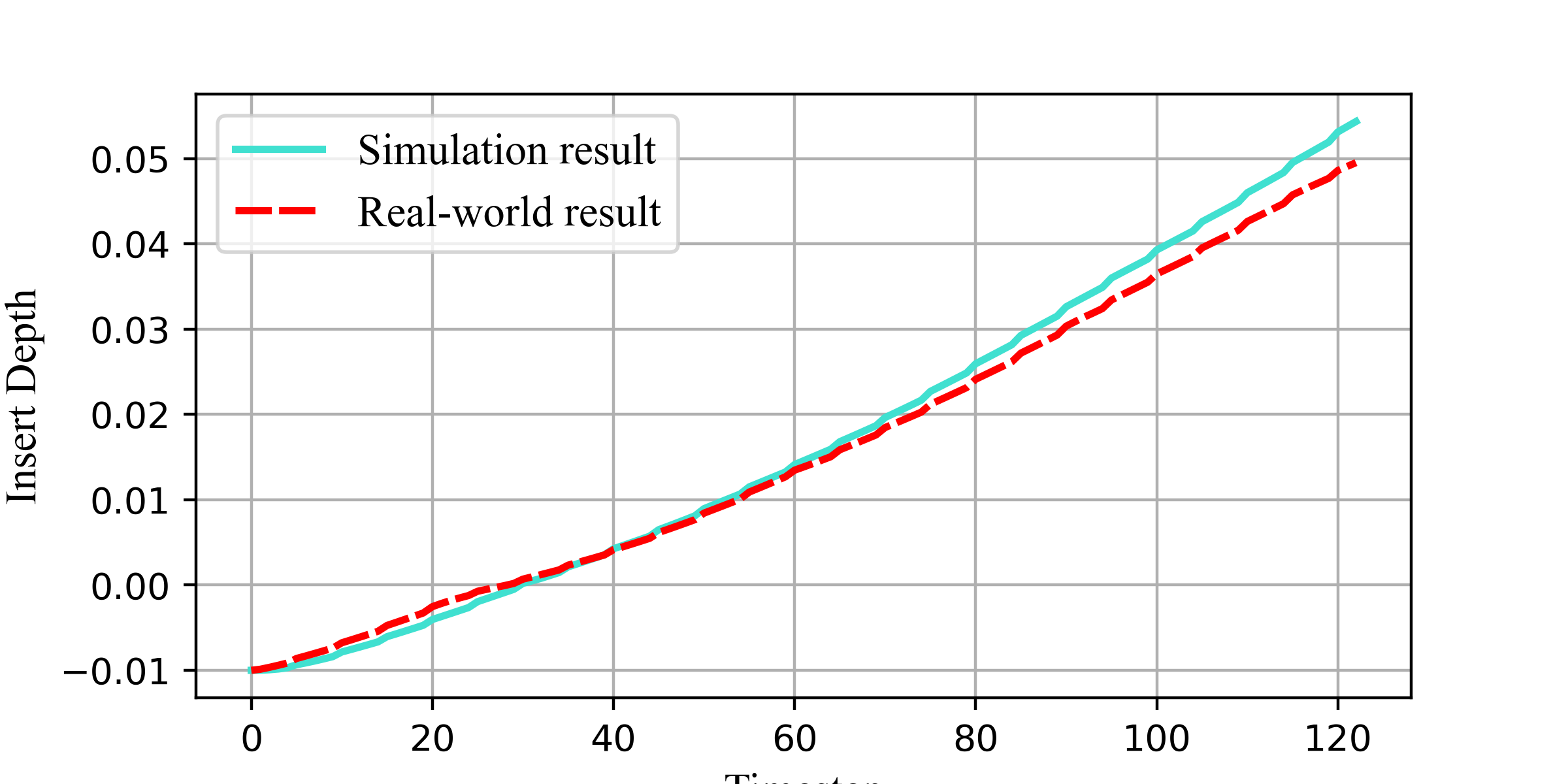}
  \caption{The peg insertion depth during the assembly in the simulation environment and real-world experiment.}
  \label{PegInsert}
\end{figure}

\begin{figure}[!t]
  \centering
  \includegraphics[width=\hsize]{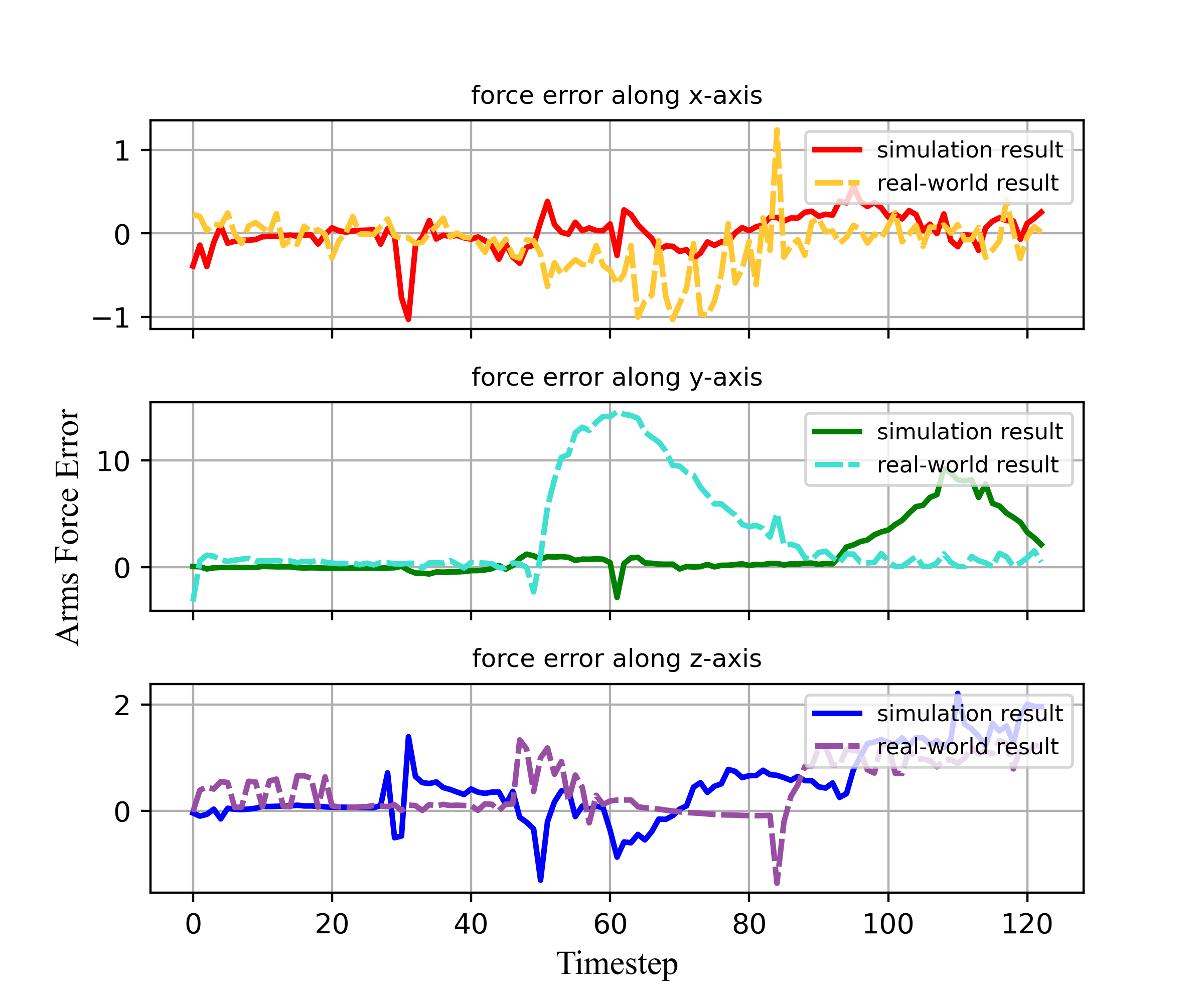}
  \caption{The contact force error bewteen the peg and the hole during the assembly in the simulation environment and real-world experiment.}
  \label{PegInsert2}
\end{figure}

After training, we test the training results on the same scenario. 
The cyan line in Fig. \ref{PegInsert} shows the peg insertion depth during the assembly in the simulation environment, and the noted lines in Fig .\ref{PegInsert2} show the contact force error bewteen the peg and the hole in the same scenario. The plot indicates that collision occurs between the peg and the hole when $timestep=30$. However, with the outer impedance control, the contact force on the X-axis (the impact direction) rapidly decreases, and the peg continues to insert into the hole. The contact forces on the X and Z axes remain within a small range throughout the insertion. In the final insertion stage, the force on the Y-axis notably increases at $timestep=90$, but it decreases from $timestep=110$ with the adjustment of \textbf{LAC}. The peg eventually reaches the target insertion depth at $timestep=123$. 

\subsubsection{Evaluation of the Generalization}
We design two test scenarios in the simulation environment to verify the generalization of our algorithm in dual-arm cooperative peg-in-hole assembly. Given the poor performance of \textbf{TD3} and \textbf{SAC}, we only compare our algorithm with the other two algorithms, named \textbf{LAC w/ fixed Imp} and \textbf{LAC w/o LSTM}.

\begin{figure}[!t]
  \centering
  \includegraphics[width=0.9\hsize]{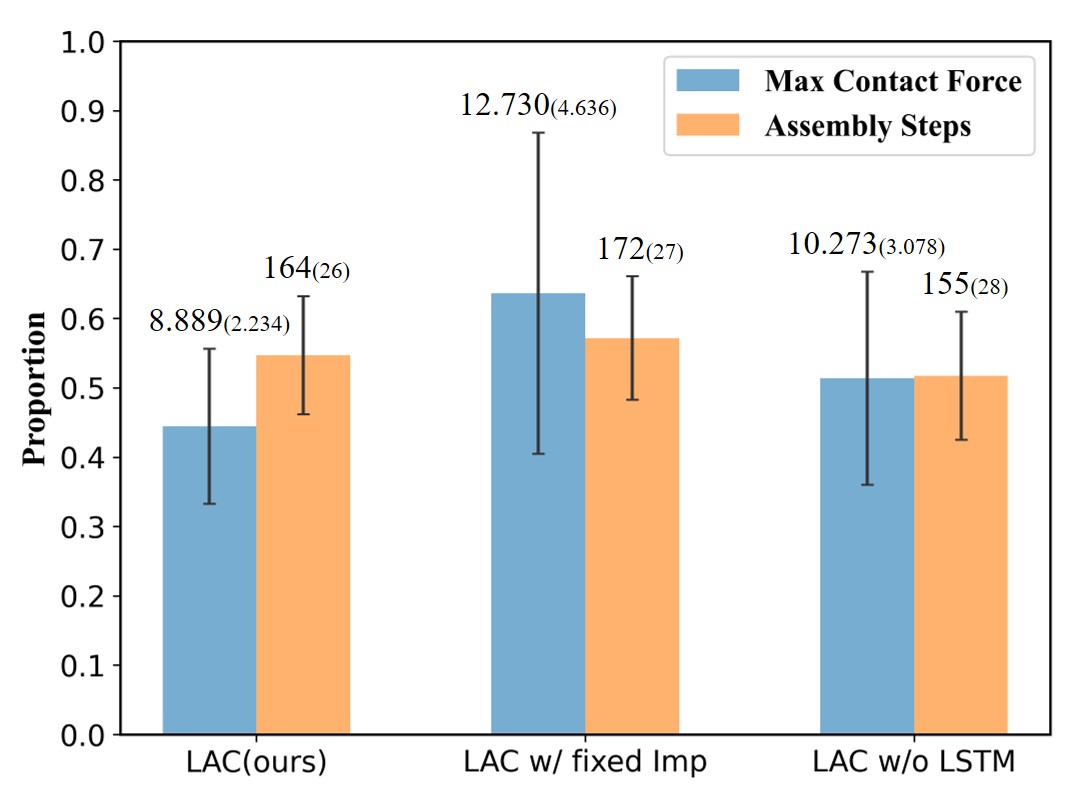}
  \caption{Mean and standard deviation of the maximum environmental contact force and assembly steps in the test environment with a random error ranging from 1mm to 5mm of the hole centre on the y-z plane. 
  }
  \label{PegErr}
\end{figure}

The first scenario is to verify the effectiveness of \textbf{LAC} in the presence of a position error in the assembly environment. Specifically, a random error ranging from 1mm to 5mm is added to the hole centre on the y-z plane. 
In 20 test experiments, our algorithm has the highest number of successful attempts (18), followed by \textbf{LAC w/o LSTM} (14), and \textbf{LAC w/ fixed Imp} with the minimum (13). Fig. \ref{PegErr} visually shows the mean and standard deviation of the three algorithms on the maximum environmental contact force and assembly steps. 
The results indicate that our method better balances assembly speed and contact force. Although the average assembly time of \textbf{LAC w/o LSTM} is slightly lower than ours, the average contact force it generated is larger. On the other hand, the average maximum contact force and assembly time of \textbf{LAC w/ fixed Imp} are the largest because it cannot adjust impedance parameters.

The other scenario is to test the adaptability of our algorithm in various material environments. 
Table \ref{PegForce} and Table \ref{PegStep} display the maximum environmental contact force and assembly steps of different algorithms in varying stiffness and damping test environments. The tables show that the variable impedance has excellent adaptability in large stiffness environments. In comparison, the maximum contact force of \textbf{LAC w/ fixed Imp} exceeds the specified safety threshold and cannot complete assembly operation in large stiffness environments. All three algorithms show an increase in maximum contact force in environments with large damping. Our algorithm's assembly steps demonstrate remarkable stability in different environments, while the assembly steps of the other two algorithms increase to varying degrees.

\begin{table}[t]
\begin{center}
\caption{Max contact force of different algorithms in different stiffness and damping test environments.}
\label{PegForce}
\begin{tabular}{|c|c|c|c|c|}
\hline
Algorithms & $K_l, B_l$/N & $K_l, B_s$/N & $K_s, B_l$/N & $K_s, B_s$/N\\
\hline
\textbf{LAC(ours)} & \textbf{12.776}  & \textbf{5.454} & \textbf{13.668} & \textbf{3.260}\\
LAC w/ fixed Imp & 27.366 & 21.812 & 13.962 & 8.815\\
LAC w/o LSTM & 14.447 & 7.063 & 14.620 & 7.597 \\
\hline 
\end{tabular}
\end{center}
\end{table}

\begin{table}[!t]
\begin{center}
\caption{Assembly steps of different algorithms in different stiffness and damping test environments.}
\label{PegStep}
\begin{tabular}{|c|c|c|c|c|}
\hline
Algorithms & $K_l, B_l$ & $K_l, B_s$ & $K_s, B_l$ & $K_s, B_s$\\
\hline
\textbf{LAC(ours)} & \textbf{124}  & \textbf{128} & \textbf{127} & \textbf{123}\\
LAC w/ fixed Imp & None & None & 201 & 127\\
LAC w/o LSTM & 142 & 138 & 147 & 130 \\
\hline 
\end{tabular}
\end{center}
\end{table}

\subsubsection{Real-World Experiments}
\begin{figure}[!t]
  \centering
  \includegraphics[width=\hsize]{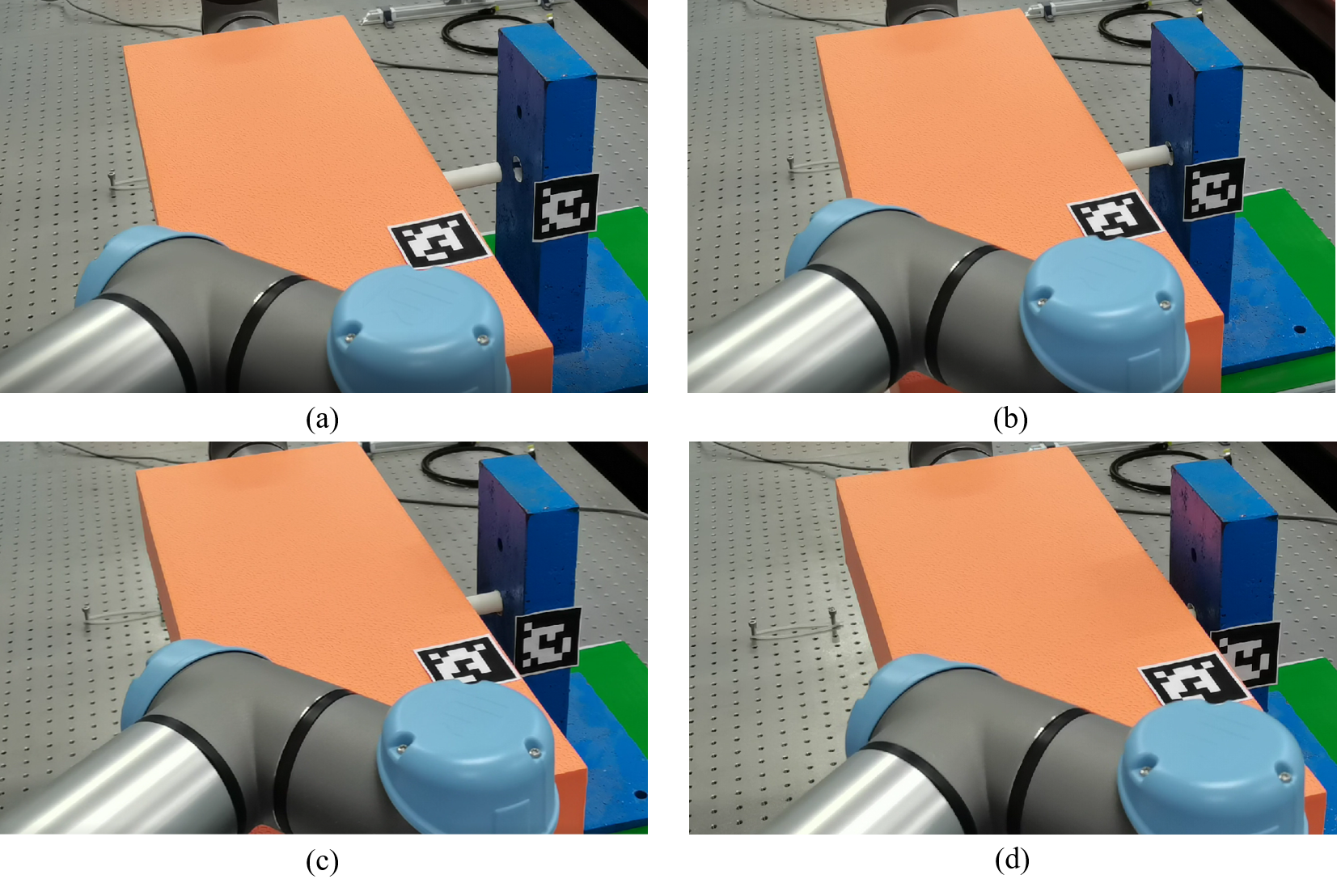}
  \caption{Real-World dual-arm assembly task. (a) Initial state of the system. (b) (c) The shaft is being inserted in the assembly hole. (d) The shaft achieves the target depth.}
  \label{rw3}
\end{figure}

We conduct the real-world peg-in-hole experiment based on the dual-arm collaborative handling operation. 
Among 20 shaft hole assembly testing experiments, the \textbf{LAC} algorithm succeeded the task 17 times totally. In the successful experiments, the average maximum contact force of the target object was 19.718N, and the average assembly time was 141 time steps (7.05s). The screenshots of the physical experiment process of the real-world dual-arm collaborative peg-in-hole task are shown in Fig. \ref{rw3}. The red line in Fig. \ref{PegInsert} shows the insertion depth of the shaft, and the noted lines in Fig. \ref{PegInsert2} show the interaction force during the assembly process. Through the contrast of the simulation results and real-world experiment results under exactly the same parameter settings, it is obvious that the corresponding lines are highly similar, which indicates that our proposed method can be easily generalized to real-world environments with high precision and robustness. In Fig. \ref{PegInsert2}, the assembly shaft and the hole come into contact at $timestamp=47 (t=2.35s)$, and the contact force along the Y-axis increases during the insertion process. However, with the adjustment of the outer loop impedance controller, this error gradually decreases from $timestamp=62 (t=3.1s)$ and eventually fluctuates around zero. 
Throughout the insertion process, the contact force along the X-axis and Z-axis remains within a small range. The insertion depth of the assembly shaft reaches the target depth at $timestep=132 (t=6.6s)$. 
The experimental results show that our proposed algorithm can achieve high-precision dual-arm assembly task within a safe range of interaction forces.

\section{CONCLUSUIN}
The current framework for symmetric bi-manual manipulation performs motion planning first and then tracks the desired trajectory with a compliance controller, which is inefficient and may cause conflicts between the two modules. Furthermore, due to the difficulty in establishing contact dynamics and the high requirements for dual-arm synchronous control, no research has particularly focused on the dual-arm cooperative peg-in-hole operation. 
To address these challenges, we propose the Learning-based Adaptive Compliance (LAC) algorithm, which incorporates the advantages of reinforcement learning to improve the flexibility, adaptability, and robustness in symmetric bi-manual manipulation. Simulation results demonstrate that the proposed algorithm 
facilitates dual-arm cooperative control with considerable environmental adaptability in high-precision assembly operations. Real-world experiments further verify the proposed method and show the high robustness and strong generalization ability of the algorithm. 

\bibliographystyle{unsrt}  

\bibliography{main}

\end{document}